\documentclass[sigconf,nonacm,screen]{acmart}
\usepackage{booktabs}
\usepackage{amsmath}
\usepackage{tikz}
\usepackage{algorithm}
\usepackage{algpseudocode}
\usetikzlibrary{positioning,fit,backgrounds,arrows.meta}

\usepackage{booktabs}
\usepackage{graphicx}
\usepackage[table]{xcolor}
\usepackage{array}
\usepackage{multirow}
\usepackage{enumitem}
\usepackage{todonotes}

\definecolor{tsblue}{HTML}{EAF2FF}
\definecolor{lobteal}{HTML}{E8F7F6}
\definecolor{tabgold}{HTML}{FFF6DF}
\definecolor{structviolet}{HTML}{F2ECFF}

\definecolor{finBlue}{HTML}{2F6F9F}
\definecolor{finTeal}{HTML}{2A9D8F}
\definecolor{finGold}{HTML}{C49A27}
\definecolor{finRed}{HTML}{B55A4A}
\definecolor{finViolet}{HTML}{6D5BA6}
\definecolor{finInk}{HTML}{263238}
\definecolor{finMist}{HTML}{EEF3F6}

\begin{document}

\title{FlowLOB: Efficient and Controllable Limit Order Book Generation with Flow Matching}

% %% Anonymous review copy. Replace with final author metadata after acceptance.
% \author{Anonymous Author(s)}
% \affiliation{%
%   \institution{Anonymous Institution(s)}
%   \country{}}

% \renewcommand{\shortauthors}{Anonymous Author(s)}

\renewcommand{\thefootnote}{\fnsymbol{footnote}}

\author{Zhuohan Wang}
\authornote{Joint affiliation with King's College London.}
\affiliation{%
  \institution{Simudyne}
  \country{United Kingdom}}
% \email{zhuohan.wang@simudyne.com}

\author{Andreea Bacalum}
\affiliation{%
  \institution{Simudyne}
  \country{United Kingdom}}
% \email{andreea@simudyne.com}

\author{Ollie Olby}
\affiliation{%
  \institution{Simudyne}
  \country{United Kingdom}}
% % \email{ollie@simudyne.com}

\author{Carmine Ventre}
\affiliation{%
  \institution{King's College London}
  \country{United Kingdom}}
% \email{carmine.ventre@kcl.ac.uk}

\author{Namid Stillman}
\affiliation{%
  \institution{Simudyne}
  \country{United Kingdom}}
\email{namid@simudyne.com}

\renewcommand{\shortauthors}{Wang et al.}

\begin{abstract}
Limit order book (LOB) simulators are most useful to practitioners when they combine realistic market dynamics, computationally efficient sampling, controllable scenario generation, and the ability to generalize beyond the instruments seen during training---properties that existing agent-based and deep generative simulators provide only partially. We present \textbf{FlowLOB}, a conditional \textbf{flow}-matching generator of \textbf{LOB} trajectories, trained on multiple Hong Kong Exchange (HKEX) symbols at three sampling frequencies ($0.1$s, $1$s, $10$s) in tick-relative representation that transfers to unseen instruments. Because flow and diffusion models admit a common formulation, we train both with identical data, architecture, and budget, and sample both through the same fixed-step ODE solvers, yielding a controlled comparison of sampling efficiency and fidelity. Flow matching attains its best quality with only $10$ ODE-solver steps, whereas diffusion needs many more function evaluations to approach the same fidelity. At this efficient operating point, FlowLOB improves realism over baselines, two learned and two agent-based models, in most distributional metrics at the two finer sampling frequencies. We evaluate counterfactual controllability with a distributional test that asks whether changing a scenario condition moves the generated statistic toward the corresponding real tail regime; FlowLOB satisfies this criterion in most tested settings. Both realism and control effects transfer zero-shot on a held-out symbol. We additionally conduct ablation studies on the network architecture and the learning rate.
\end{abstract}


\maketitle

\section{Introduction}
\label{sec:intro}

Simulation is central to many workflows in quantitative finance. Backtesting a trading strategy, stress-testing an execution algorithm, and training a reinforcement-learning agent all require far more market data than history provides: for any instrument, the market offers exactly one realized path~\cite{potluru2023synthetic,buehler2019deephedging}. There are no repeated trials and no counterfactuals---no way to observe how the same trading day would have unfolded under higher volatility or thinner liquidity. A simulator of the limit order book (LOB), the fundamental data structure of modern electronic markets, is the standard escape from this constraint~\cite{vyetrenko2020getreal,byrd2020abides}, and its value hinges on three properties: the realism of what it generates, the cost of generating it, and the degree to which it can be steered toward scenarios of interest.

% Illustrate problems we are faced with.
% 1. Fast sampling
% 2. Multi-symbol generalization
% 3. Controllablity
Existing LOB simulators deliver these properties only in part. Agent-based models---zero-intelligence traders~\cite{gode1993allocative, farmer2005predictive} and Hawkes-process order flow~\cite{bacry2015hawkes}, composed into platforms such as ABIDES~\cite{byrd2020abides}---are interpretable and easy to steer, but reproduce only a subset of the stylized facts of real markets~\cite{cont2001empirical}. Deep generative models close the realism gap: generative adversarial networks~\cite{wiese2020quantgan,li2020generating,coletta2021towards, coletta2022learning}, autoregressive event models~\cite{nagy2023generative,li2025bytegen}, and most recently diffusion models~\cite{berti2025trades,backhouse2025painting,zheng2024lobdif, wang2025diffvolume,wang2026difflob} produce order books that are difficult to distinguish from real data. But they come at a price. First, diffusion sampling is expensive by construction~\cite{ho2020ddpm,song2021sde}: a single trading day at $0.1$-second resolution comprises roughly $200{,}000$ book states, so a $50$-step sampler spends $50$ network evaluations on each. Fast samplers~\cite{songddim2021,lu2022dpm, zhao2023unipc,song2023consistency} mitigate this, but treat the symptom rather than the problem. Second, these models are almost universally trained one-model-per-symbol, so nothing can be said about instruments outside the training set. Third, while conditional generation is common, conditional \emph{validity} is rarely tested: models are conditioned on a regime, but whether the generated data actually belong to that regime is left unmeasured. 
What practitioners need is a LOB generator that is simultaneously realistic, cheap enough to sample at scale, controllable with measurable effect, and general across instruments. No existing simulator provides all four.

We present \textbf{FlowLOB}, a flow-matching~\cite{lipman2023flow, liu2023rectifiedflow} generative model of limit order book trajectories. FlowLOB models windows of book states (ten price levels per side, prices and volumes) with a $\sim$100M-parameter adaLN-Zero transformer~\cite{peebles2022dit}, trained jointly on eight Hong Kong Exchange symbols at three different sampling frequencies ($0.1$s, $1$s, $10$s), and conditioned on four interpretable channels: trend, volatility, liquidity, and order-book imbalance. 

Our contributions are fourfold:

\begin{itemize}[topsep=5pt,itemsep=0.5ex]
  \item \textbf{Generality.} For each sampling frequency, FlowLOB is trained as a shared simulator across eight HKEX symbols rather than one model per symbol. The tick-relative representation enables zero-shot transfer to a ninth held-out symbol without retraining, and we repeat the evaluation at $0.1$s, $1$s, and $10$s to test this behavior across temporal resolutions.
  \item \textbf{Efficient sampling.} We place flow models and diffusion models on equal footing: one dataset, one backbone, one training recipe, and one family of fixed-step ODE solvers~\cite{karras2022edm}. Under this matched setup, 10 Euler steps already put FlowLOB at its best or nearly best quality. At the same sampling compute, its marginal distributional error is $4$--$120\times$ lower than diffusion across price and volume features.
  \item \textbf{Realism.} We compare the pooled price and volume distributions of generated books against real data under four distance metrics, against two learned baselines (LOB-S5~\cite{nagy2023generative}, LOB-GAN~\cite{coletta2021towards}) and two agent-based baselines (Hawkes~\cite{bacry2015hawkes, jain2024limit}, Zero Intelligence traders~\cite{gode1993allocative, farmer2005predictive}). FlowLOB attains the smallest distance in $30$ of the $32$ resulting parameter combinations at the $0.1$s and $1$s frequencies, on both the in-distribution and the held-out symbol.
  \item \textbf{Controllability.} Conditioning passes a distributional validity test, moving the generated statistic's distribution toward the real tail regime in $42$ of $48$ cases considered, with liquidity and imbalance reliably steerable at every frequency.
\end{itemize}
% \todo{add a short paragraph here to say how close flowlob is to what practitioners need.}

From a practitioner perspective, FlowLOB matches key requirements for synthetic market-data engines used in downstream ML development and validation \cite{nie2026deep, olby2025right}. It samples efficiently enough for large-scale rollouts, steers rare regimes, transfers across instruments including unseen ones, and outputs L2 book states suitable for snapshot-based policies such as dynamic execution.

\section{Related Work}
\label{sec:related}

\noindent \textbf{Realistic and controllable LOB simulation.}
Market simulation has traditionally been agent-based: zero-intelligence traders~\cite{gode1993allocative,farmer2005predictive} and Hawkes-process order flow~\cite{bacry2015hawkes}, composed into platforms such as ABIDES~\cite{byrd2020abides}. These models are transparent and steerable, but realism audits find they reproduce only part of the stylized facts of
real markets~\cite{cont2001empirical,vyetrenko2020getreal}. Learned generators progressively closed this gap: GANs over order streams and
book states~\cite{li2020generating,coletta2021towards,coletta2022learning, hultin2023generative}, autoregressive event models~\cite{nagy2023generative,li2025bytegen}, and most recently diffusion models, which set the current realism standard for both message-level and book-level generation~\cite{berti2025trades,backhouse2025painting,zheng2024lobdif, wang2025diffvolume}. The controllability that agent-based models offered has been rebuilt on top of these learned generators only recently, typically via conditional or guided diffusion~\cite{ho2022cfg}: DIGMA~\cite{huang2024digma} guides a diffusion-based meta agent toward target market dynamics, CoFinDiff~\cite{tanaka2025cofindiff} conditions financial time series on trend and volatility, and DiffLOB~\cite{wang2026difflob} combines a WaveNet-style diffusion backbone with ControlNet-style conditioning to generate counterfactual order books under high and low regimes of book-level statistics.
Following DiffLOB, we focus on L2 order book states generation rather than order flow generation. 

\noindent \textbf{Diffusion, flow matching, and efficient sampling.}
Diffusion models generate by learning to invert a gradual noising process~\cite{sohldickstein2015,ho2020ddpm}, a view unified in continuous time by score-based SDEs~\cite{song2019ncsn,song2021sde}, whose deterministic probability-flow ODE permits sampling by numerical integration. Paired with transformer backbones~\cite{peebles2022dit} and guidance~\cite{dhariwal2021beatgan, ho2022cfg}, this family dominates generative modeling across domains~\cite{yang2023diffusion}, including the financial applications above. Its main practical liability is inference cost: high-quality samples require many network evaluations along the reverse process~\cite{ho2020ddpm,song2021sde}. One line of work accelerates a trained diffusion model post hoc, through implicit samplers~\cite{songddim2021}, high-order and dedicated ODE solvers~\cite{karras2022edm,lu2022dpm,zhao2023unipc}, and distillation into consistency models~\cite{song2023consistency}. A second line changes the training objective itself: flow matching~\cite{lipman2023flow} and rectified flow~\cite{liu2023rectifiedflow} learn near-straight probability paths that are cheap to integrate by construction, are unified with diffusion under the stochastic-interpolant view~\cite{albergo2023interpolants,holderrieth2026introflowdiffusion}, and scale to state-of-the-art generation~\cite{esser2024sd3}. In finance, flow matching has reached time-series generation~\cite{hu2024flowts,he2025timeflow} and trading policies~\cite{li2025flowhft}, but not the order book itself, and no prior work compares the two objectives for market generation under matched data, architecture, and compute. 

% Should add more inference in this section
\section{Methodology}
\label{sec:method}

The methodology section has four parts: a matched flow/diffusion training setup, a common ODE sampling protocol, a tick-relative representation for multi-symbol LOB data, and a transformer backbone that aligns market history with scenario controls.

\subsection{Flow and Diffusion Training}

Let $x_0 \sim p_{\mathrm{data}}(x \mid c)$ denote a target LOB window and let $c$ denote the conditioning tensor constructed from the preceding market state and scenario variables. We train two conditional generative objectives on the same pairs $(x_0,c)$: flow matching~\cite{lipman2023flow, liu2023rectifiedflow} and diffusion~\cite{ho2020ddpm, song2021sde}. This pairing lets us compare two ways of learning a transport from noise to realistic book trajectories while holding the data, conditioning, architecture, and optimizer fixed.

Flow matching learns a velocity field along a prescribed path from Gaussian noise to data. With $z \sim \mathcal{N}(0,I)$ and linear interpolation $x_t=(1-t)z+t x_0$ for $t\in[0,1]$, the target velocity is constant, $x_0-z$. The flow model $v_\theta(x_t,c,t)$ is trained by
\[
    \mathcal{L}_{\mathrm{FM}}(\theta)
    =
    \mathbb{E}_{t,z,(x_0,c)}
    \left[
        \left\|v_\theta(x_t,c,t) - (x_0-z)\right\|_2^2
    \right].
\]
Sampling then follows the learned velocity field from noise at $t=0$ to data at $t=1$.

The diffusion baseline is trained in the same conditional regression template, but uses the variance-preserving noising path $x_t=\alpha_t x_0+\sigma_t \epsilon$, with $\epsilon\sim\mathcal{N}(0,I)$ and schedule
$(\alpha_t,\sigma_t)$ moving data toward Gaussian noise as $t$ increases. The network predicts the injected noise,
\[
    \mathcal{L}_{\mathrm{DM}}(\theta)
    =
    \mathbb{E}_{t,\epsilon,(x_0,c)}
    \left[
        \left\|\epsilon_\theta(x_t,c,t)-\epsilon\right\|_2^2
    \right].
\]
At sampling time this prediction is converted into the probability-flow ODE
field used in Sec.~\ref{sec:sampling}.

\subsection{ODE Sampling and Inference Cost}
\label{sec:sampling}

% Both trained objectives are sampled through deterministic ODEs, which gives a common way to measure generation cost. For flow matching, the learned velocity is used directly, $u(x,t)=v_\theta(x,c,t)$, and the ODE is integrated from $t=0$ to $t=1$. For diffusion, the noise prediction is converted into the corresponding probability-flow ODE field and integrated in the reverse time direction, from the noisy endpoint toward the data endpoint. In both cases, the conditioning tensor $c$ is held fixed throughout the trajectory.

Both trained objectives are sampled through deterministic ODEs, which gives a common way to measure generation cost. In both cases, sampling can be written as integrating a learned field
\[
    \frac{d x_t}{dt} = u(x_t,t,c),
\]
with the conditioning tensor $c$ held fixed throughout the trajectory. For flow matching, the learned velocity is used directly,
$u(x,t,c)=v_\theta(x,c,t)$, and the ODE is integrated from $t=0$ to $t=1$.
For diffusion, the noise prediction is converted into the corresponding probability-flow ODE field and integrated in the reverse time direction, from the noisy endpoint toward the data endpoint.

We sample both models with the same family of fixed-step solvers: Euler, Heun, and fourth-order Runge--Kutta (RK4). These solvers require one, two, and four network evaluations per step, respectively, so a run with $N$ solver steps costs $N$, $2N$, or $4N$ neural function evaluations (NFE). This accounting lets us compare solvers at equal model-call budgets rather than equal step counts. We use NFE, rather than wall-clock time, as the primary inference-cost unit because it isolates the algorithmic cost of sampling from implementation details such as batching, hardware, and kernel efficiency.

This shared sampler is what makes the flow-versus-diffusion comparison
controlled. Each sweep varies only the numerical method and step count applied to a fixed checkpoint; the training data, conditioning variables, backbone, and optimizer are unchanged. Thus any quality--cost difference observed in Sec.~\ref{sec:exp-nfe} reflects the learned generative dynamics and their ease of numerical integration, rather than a different architecture or sampling budget.

% We should mention how we adapt processing methods for HK data because of differnet tick sizes
\subsection{LOB Representation and Conditioning}
\label{sec:data-tensors}

The raw order book is first converted into fixed-length windows on a uniform time grid. At each sampling frequency, we resample the top ten ask and bid levels by taking the first observation in each bin and forward-filling empty bins. A training example consists of two consecutive windows: a past window, which provides the observed market state, and a target window. The generated target is represented as $x\in\mathbb{R}^{2\times 20\times 32}$, with two channels for price and volume, twenty book levels ordered from the outer ask to the outer bid, and thirty-two future time steps.

Prices are not represented in absolute currency units. This is important for HKEX data because different symbols can trade at different price levels and under different tick sizes. We therefore use a tick-relative local representation: the first price coordinate records the mid-price change between consecutive states, while the remaining coordinates record adjacent level gaps within the book, all measured in ticks. Volumes $V$ are transformed level-wise as $\log(V+1)$. Together, tick normalization for prices and logarithmic normalization for volumes remove much of the instrument-specific scale while preserving the local geometry and depth profile of the book. This makes the pooled multi-symbol training problem better posed and is a key reason the same model can be evaluated zero-shot on a held-out symbol.

The conditioning tensor $c\in\mathbb{R}^{7\times 20\times 32}$ is aligned with the target window. It contains the past-window price and volume representation, time of day and four scenario channels. 
We include time of day to capture HKEX's strong intraday structure, including the midday lunch break and the double-U-shaped volume profile. Unlike DiffLOB~\cite{wang2026difflob}, which discards the first and last trading hours, we retain the full continuous-auction period. The computed time-of-day ratio is clamped to $[0,1]$ at session boundaries.
The scenario channels encode trend, volatility, liquidity, and order-book imbalance. Trend and volatility are window-level summaries and are broadcast across levels and time, while liquidity and imbalance are computed at each future time step and broadcast across levels. This construction gives the model both the recent market state and an explicit description of the regime the generated continuation should satisfy.

\subsection{Architecture}

FlowLOB uses an adaLN-Zero transformer~\cite{peebles2022dit,perez2018film}
backbone. We choose this backbone because adaptive normalization has become a standard scalable way to condition large generative transformers, and because it separates global process-time modulation from local market conditioning. Each cell of the level--time grid is treated as one token, so a target window is represented as $20\times32$ tokens. The noisy state $x_t$ and conditioning tensor $c$ are concatenated channel-wise before projection, which preserves the alignment between generated quantities and their controls: a liquidity or imbalance value specified for a particular future step is attached to the tokens for that same step.

We use separate learned embeddings for book level and time step, and inject the ODE time $t$ through adaptive layer-normalization blocks. Thus market conditioning enters through the token features, while the generative process time enters through modulation of the transformer dynamics. A linear output head maps the final token representations back to the price and volume channels of the generated LOB window.

\section{Experimental Results}
\label{sec:experiments}

% Our evaluation follows the dependency structure of the paper's claims.
% Sec.~\ref{sec:exp-setup} fixes data, training settings, baselines, and
% metrics. Sec.~\ref{sec:exp-nfe} sweeps solvers and step counts to locate
% the quality-versus-cost operating point, which
% Secs.~\ref{sec:exp-realism}--\ref{sec:exp-ablation} then inherit: all
% subsequent comparisons share checkpoints, prediction type, data slice, and
% sampling settings, varying only the factor under study.

In this section, we first specify the data, model settings, and baselines. We then compare flow matching and diffusion under matched samplers to choose a quality--cost operating point (Sec.~\ref{sec:exp-nfe}). Using that fixed sampling budget, we evaluate realism against learned and agent-based baselines on both an in-distribution symbol and a held-out symbol (Sec.~\ref{sec:exp-realism}). Finally, we test whether the scenario variables defined in Sec.~\ref{sec:data-tensors} produce measurable counterfactual changes (Sec.~\ref{sec:exp-counterfactual}) and ablate the main architectural and optimization choices (Sec.~\ref{sec:exp-ablation}). Unless stated otherwise, each experiment reuses the same trained checkpoints and varies only the factor under study.

\subsection{Experimental Setup}
\label{sec:exp-setup}

\noindent \textbf{Data.}
We use level-2 order book data from the Hong Kong Exchange (HKEX), ten price levels per side, for eight symbols spanning technology, financials, automotive, and semiconductors, i.e., 5, 700, 981, 1024, 1211, 1299, 1810, and 9618 (all .HK). Trading sessions are 09:30--12:00 and 13:00--16:00, giving $5.5$ trading hours per day---$198{,}000$, $19{,}800$, and $1{,}980$ grid points at the $0.1$s, $1$s, and $10$s frequencies respectively. Training data covers 2025-09-01 to 2025-11-15, validation 2025-11-16 to 2025-11-30, and all evaluation uses December 2025, which covers a variety of market conditions due to the end of year period. One further symbol, 9999.HK, is excluded from training entirely and evaluated zero-shot. Throughout, we report results on 700.HK as a representative in-distribution symbol and 9999.HK as the zero-shot held-out symbol; results are qualitatively similar across the other in-distribution symbols.

% \paragraph{Training settings.}
\noindent \textbf{Training settings.}
All runs share the remaining configuration: AdamW without weight decay, effective batch size $1024$ (micro-batch $256$, $4$ accumulation steps), up to $500$ epochs with early stopping (patience $20$), and conditioning dropout $0.5$ for classifier-free guidance. Both transformer and UNet settings use the $\sim$100M configuration. For every run, we evaluate the checkpoint with the best validation loss.

% \paragraph{Baselines.}
\noindent \textbf{Baselines.} We compare against both agent-based and learned LOB generators. The zero-intelligence (ZI) and compound Hawkes (reported as Hawkes) baselines are adapted from the calibration procedures of Kawawa-Beaudan et al.~\cite{kawawa2026tradefm}. ZI samples order side, action type, inter-arrival time, volume, and price depth from fitted marginal distributions, providing a stochastic baseline that reproduces simple empirical marginals but not conditional market dynamics. Hawkes adds temporal dependence by fitting a multivariate Hawkes process to event arrivals grouped by action and side, while sampling volumes and price depths from fitted marginal distributions. We also compare against learned baselines for LOB generation. LOB-GAN follows the conditional GAN architecture of Coletta et al.~\cite{coletta2021towards}, generating order flow from a real history window. LOB-S5 treats LOB generation as an autoregressive sequence-modelling problem using a structured state-space network~\cite{nagy2023generative}. All baseline outputs are converted to the same sampled LOB representation before evaluation.

\subsection{Inference Quality vs.\ Cost}
\label{sec:exp-nfe}
\begin{figure}[t]
\centering
\includegraphics[width=\linewidth]{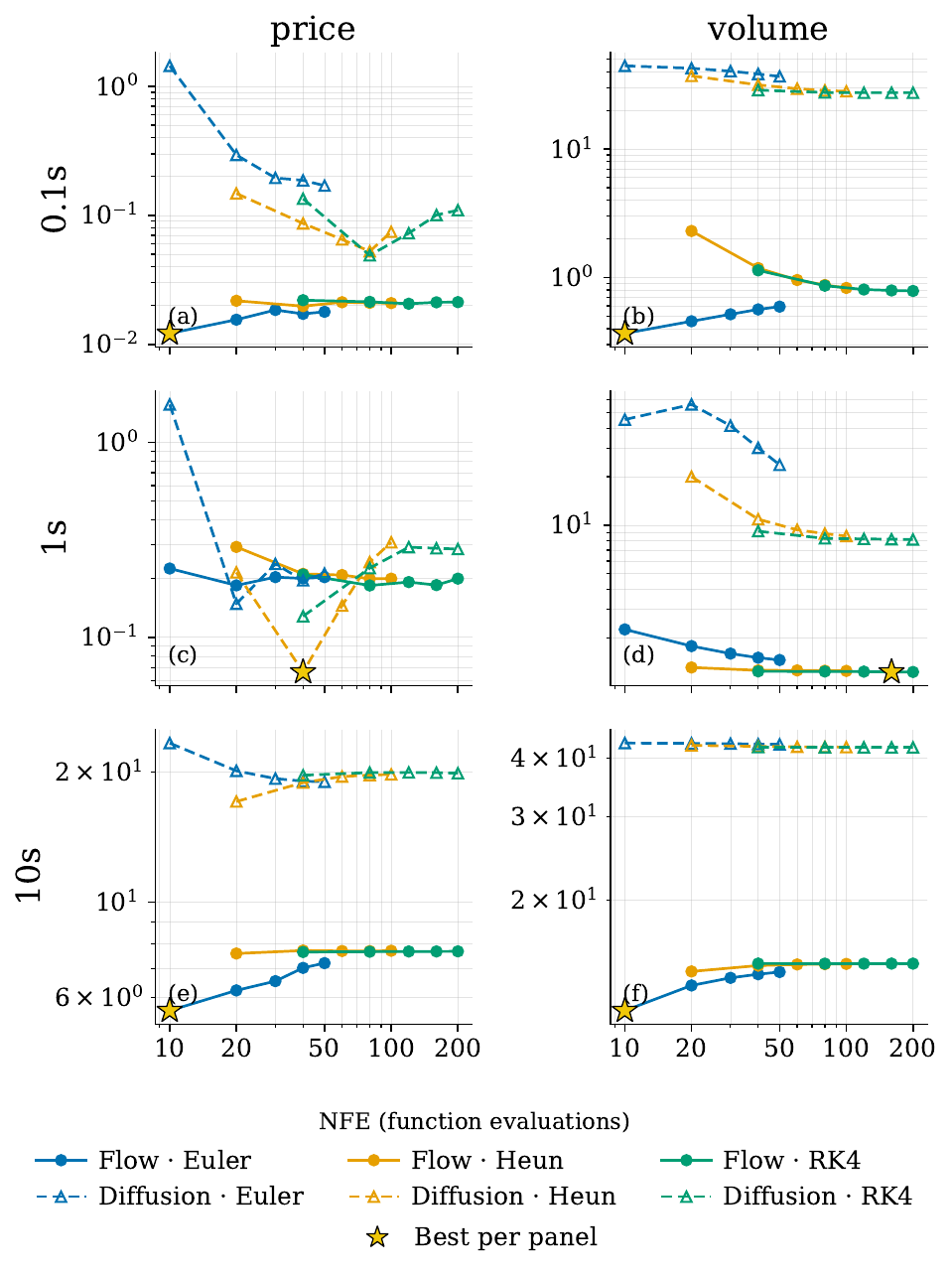}
\caption{Sample quality versus inference cost. $W_1$ distance to the real
pooled price (left) and volume (right) marginals at the three sampling
frequencies (rows), against NFE on a log--log scale. Solid: flow matching;
dashed: diffusion; one curve per solver; the star marks the best point per
panel. Flow with Euler at $N{=}10$ --- the cheapest configuration in the
sweep --- is the best point in four of six panels.}
\label{fig:quality-cost}
\end{figure}
% setting
For each sampling frequency, we take the trained flow-matching and diffusion checkpoints from Sec.~\ref{sec:exp-setup} and vary only the ODE solver and number of solver steps. We sweep Euler, Heun, and RK4 over $N\in\{10,20,30,40,50\}$, and report inference cost in NFE. 
We use Wasserstein-1 distance $W_1$ as a lightweight quality proxy to compare generated and real pooled marginals of the price and volume features.
% dissect figure
Figure~\ref{fig:quality-cost} shows that flow matching reaches its best quality at very small sampling budgets. 
With Euler sampling at $10$ steps, FlowLOB obtains lower $W_1$ than the matched diffusion sampler by $4$--$120\times$ across price and volume marginals. Even when diffusion is allowed up to $200$ NFEs, no diffusion configuration in our sweep matches FlowLOB-Euler-10 in five of the six settings. The only exception is the $1$s price marginal, where diffusion matches at $20$ NFEs by Heun; however, the same configurations remain substantially worse on volume.

% what is the cause
% 1. Why flow is better than diffusion on the same NFE
% 2. Why increasing ODE steps dont imporve performance
% 3. Why lower order ODE samplers is better than higher order ODE samplers for flow
These patterns suggest that solver error is already small for FlowLOB at low NFE. Flow matching learns a deterministic transport that coarse Euler can follow well, while diffusion must resolve a reverse denoising path across noise levels, so errors accumulate at the same NFE. Once flow solver error is small, more ODE steps mainly integrate the same imperfect vector field more accurately, which need not improve distributional quality. Higher-order solvers can also query intermediate off-path states where the learned field is less calibrated, whereas coarse Euler may smooth over local model errors.

This reduction in sampling cost has practical consequences. In applications such as portfolio risk analysis, execution stress testing, or reinforcement learning, a simulator is rarely queried once for a single instrument. Users typically need many trajectories across multiple symbols, regimes, and random seeds. In that setting, the difference between a sampler that is accurate after ten network evaluations and one that requires many tens or hundreds of evaluations directly compounds  into the number of scenarios that can be explored under a fixed compute budget.

We therefore use flow matching with Euler sampling at $N=10$ as the default operating point for the rest of the paper. This is the cheapest configuration in the sweep and is already on the observed quality--cost frontier. All subsequent FlowLOB realism and controllability experiments inherit this setting unless stated otherwise.

\begin{figure*}[!htbp]
\centering
\includegraphics[width=0.9\textwidth]{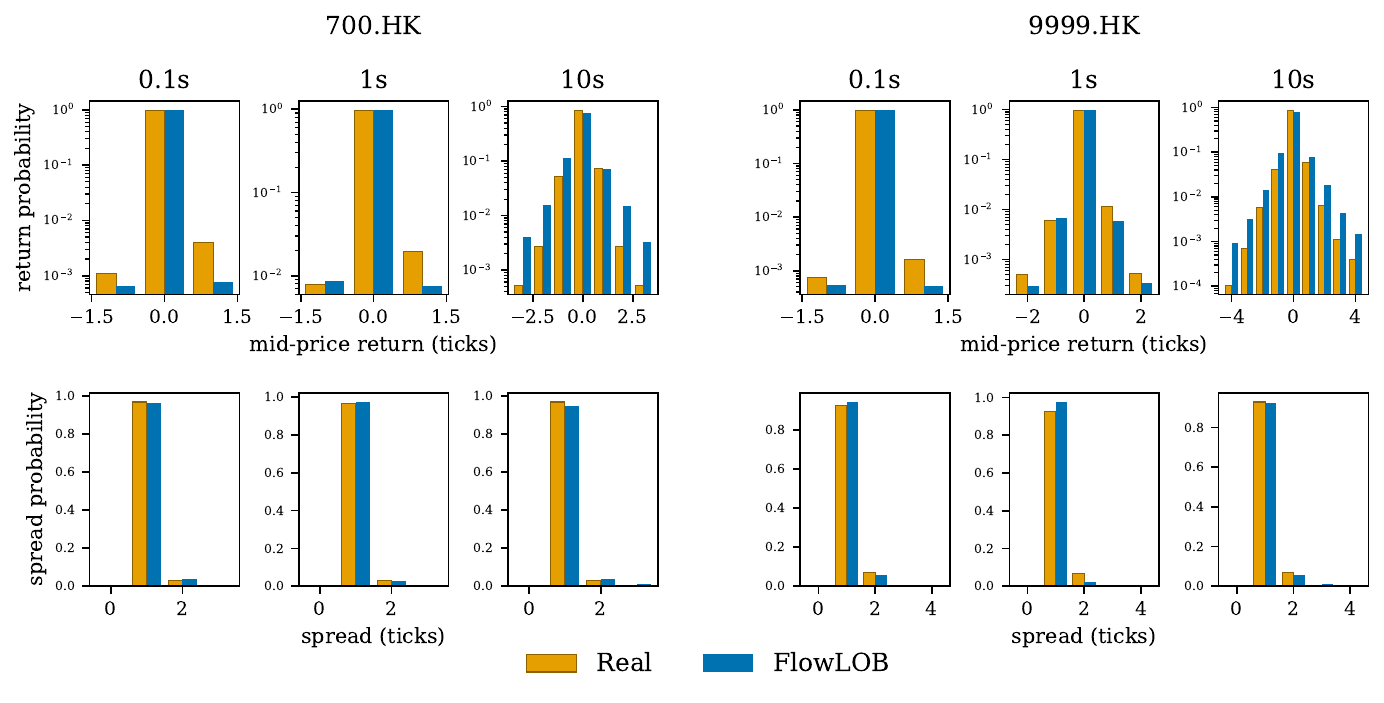}

\vspace{0.6em}

\includegraphics[width=0.9\textwidth]{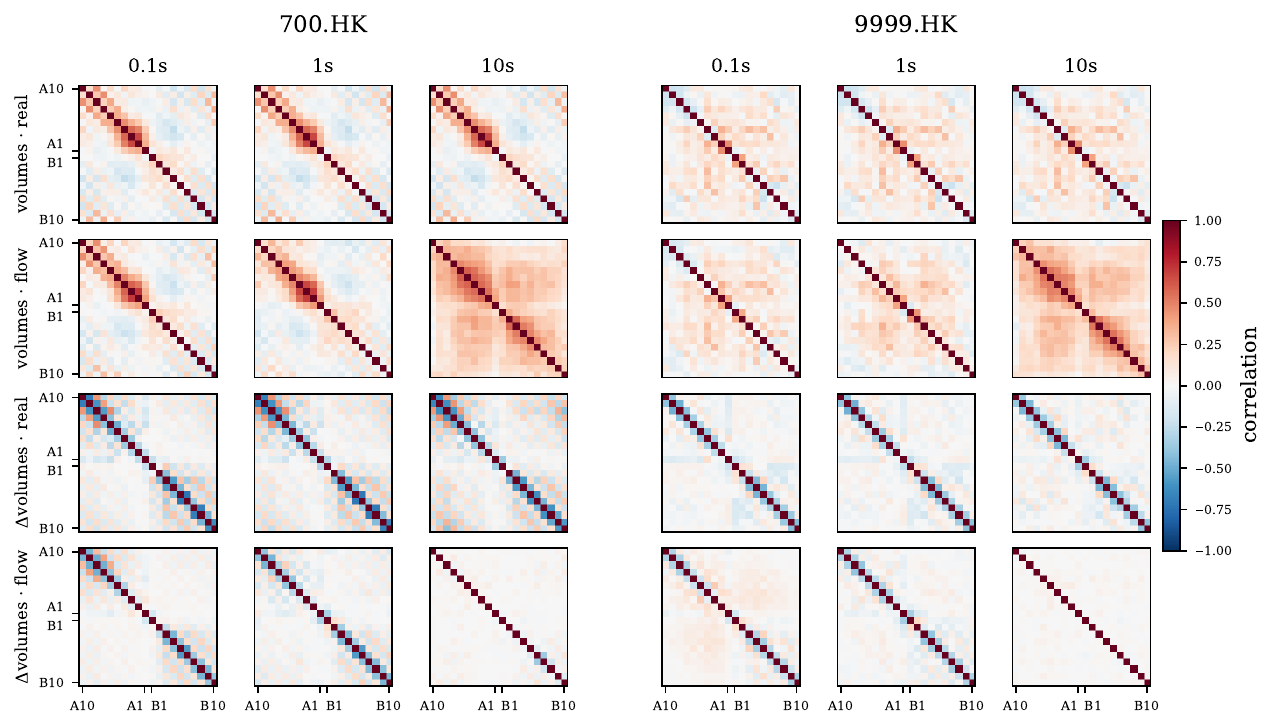}

\caption{Realism diagnostics beyond pooled marginals.
Top subplot is about Mid-price return and bid--ask spread
distributions. 
Bottom subplot is about cross-level correlation of volumes and volume changes across the 20 book levels (A10--A1, B1--B10). Real versus FlowLOB, for 700.HK and the held-out 9999.HK across the three sampling frequencies.}
\label{fig:realism-diagnostics}
\end{figure*}

\begin{table*}[t]
\caption{Realism: distances between generated and real pooled marginals
(lower is better; best per frequency--symbol block in bold).}
\label{tab:realism}
\centering
\setlength{\tabcolsep}{2.2pt}
\scalebox{0.76}{
\begin{tabular}{ll rrrr rrrr rrrr rrrr}
\toprule
& & \multicolumn{8}{c}{700.HK} & \multicolumn{8}{c}{9999.HK} \\
\cmidrule(lr){3-10}\cmidrule(lr){11-18}
& & \multicolumn{2}{c}{$W_1$} & \multicolumn{2}{c}{KS} & \multicolumn{2}{c}{KL} & \multicolumn{2}{c}{JS}
& \multicolumn{2}{c}{$W_1$} & \multicolumn{2}{c}{KS} & \multicolumn{2}{c}{KL} & \multicolumn{2}{c}{JS} \\
\cmidrule(lr){3-4}\cmidrule(lr){5-6}\cmidrule(lr){7-8}\cmidrule(lr){9-10}
\cmidrule(lr){11-12}\cmidrule(lr){13-14}\cmidrule(lr){15-16}\cmidrule(lr){17-18}
Freq. & Method & price & volume & price & volume & price & volume & price & volume & price & volume & price & volume & price & volume & price & volume \\
\midrule
\multirow{5}{*}{0.1s}
& FlowLOB & \textbf{0.01683} & \textbf{791.8} & \textbf{0.002337} & \textbf{0.01066} & \textbf{6.572e-5} & \textbf{0.01025} & \textbf{1.923e-5} & \textbf{0.002559}
& \textbf{0.008337} & \textbf{1020} & \textbf{0.01511} & \textbf{0.02853} & \textbf{2.785e-5} & \textbf{0.008375} & \textbf{6.852e-6} & \textbf{0.002289} \\
& LOB-S5 & 0.5660 & 4610 & 0.03852 & 0.04304 & 0.009084 & 0.2461 & 0.002291 & 0.008273
& 0.7296 & 3709 & 0.1392 & 0.1590 & 0.1173 & 0.1868 & 0.02119 & 0.04011 \\
& LOB-GAN & 0.8921 & 3.150e4 & 0.08873 & 0.6158 & 0.06465 & 1.216 & 0.01401 & 0.2391
& 0.7901 & 6789 & 0.1439 & 0.3306 & 0.1087 & 0.5501 & 0.02190 & 0.1198 \\
& Hawkes & 10.73 & 2.995e4 & 0.1137 & 0.5321 & 0.08033 & 1.200 & 0.02471 & 0.2284
& 9.053 & 9482 & 0.09567 & 0.2976 & 0.1336 & 0.4521 & 0.03211 & 0.1064 \\
& ZI & 0.1461 & 7337 & 0.009761 & 0.1414 & 0.003240 & 0.08502 & 0.0009035 & 0.02087
& 0.3941 & 7212 & 0.06502 & 0.1981 & 0.01688 & 0.2110 & 0.004859 & 0.05524 \\
\midrule
\multirow{5}{*}{1s}
& FlowLOB & 0.1518 & \textbf{3121} & 0.01433 & \textbf{0.02311} & \textbf{0.001220} & \textbf{0.01731} & \textbf{0.0003604} & \textbf{0.003681}
& \textbf{0.08184} & \textbf{1540} & \textbf{0.02445} & \textbf{0.07212} & \textbf{0.0006185} & \textbf{0.05187} & \textbf{0.0001562} & \textbf{0.01203} \\
& LOB-S5 & 0.5613 & 4601 & 0.03828 & 0.04296 & 0.009078 & 0.2299 & 0.002268 & 0.008303
& 0.7304 & 3703 & 0.1392 & 0.1590 & 0.1170 & 0.1854 & 0.02114 & 0.04001 \\
& LOB-GAN & 0.8904 & 3.156e4 & 0.08813 & 0.6166 & 0.06480 & 1.195 & 0.01401 & 0.2396
& 0.7933 & 6808 & 0.1445 & 0.3314 & 0.1087 & 0.5428 & 0.02194 & 0.1199 \\
& Hawkes & 10.55 & 2.995e4 & 0.1137 & 0.5325 & 0.08002 & 1.213 & 0.02456 & 0.2288
& 9.023 & 9486 & 0.09600 & 0.2980 & 0.1343 & 0.4509 & 0.03221 & 0.1065 \\
& ZI & \textbf{0.1454} & 7331 & \textbf{0.009527} & 0.1415 & 0.003173 & 0.08490 & 0.0008823 & 0.02086
& 0.3922 & 7217 & 0.06490 & 0.1988 & 0.01693 & 0.2112 & 0.004868 & 0.05536 \\
\midrule
\multirow{5}{*}{10s}
& FlowLOB & 4.942 & 1.541e4 & 0.2256 & 0.1762 & 0.2462 & 0.1931 & 0.07953 & 0.02425
& 2.113 & 3835 & 0.1548 & \textbf{0.1605} & 0.2861 & \textbf{0.1230} & 0.08507 & \textbf{0.03059} \\
& LOB-S5 & 0.5574 & \textbf{4413} & 0.03816 & \textbf{0.04252} & 0.008888 & 0.2150 & 0.002305 & \textbf{0.007948}
& 0.7181 & \textbf{3745} & 0.1368 & 0.1631 & 0.09986 & 0.1825 & 0.01986 & 0.03869 \\
& LOB-GAN & 0.8688 & 3.056e4 & 0.08414 & 0.5960 & 0.04392 & 1.172 & 0.01014 & 0.2267
& 0.7752 & 7336 & 0.1421 & 0.3470 & 0.1049 & 0.5027 & 0.02191 & 0.1186 \\
& Hawkes & 10.49 & 2.997e4 & 0.1138 & 0.5314 & 0.09067 & 1.280 & 0.02674 & 0.2281
& 8.721 & 9509 & 0.09620 & 0.2996 & 0.1251 & 0.4582 & 0.02969 & 0.1068 \\
& ZI & \textbf{0.1445} & 7250 & \textbf{0.009626} & 0.1401 & \textbf{0.002735} & \textbf{0.08137} & \textbf{0.0007391} & 0.02028
& \textbf{0.3918} & 7107 & \textbf{0.06520} & 0.1965 & \textbf{0.01484} & 0.2046 & \textbf{0.004311} & 0.05383 \\
\bottomrule
\end{tabular}}
\end{table*}

\subsection{Realism}
\label{sec:exp-realism}

% describe Figure 2
We next evaluate whether FlowLOB generates realistic book continuations at the operating point selected in Sec.~\ref{sec:exp-nfe}. Figure~\ref{fig:realism-diagnostics} first compares generated and real books through derived market statistics and cross-level dependence. The top row shows mid-price return and bid--ask spread distributions for 700.HK and the held-out 9999.HK across all three sampling frequencies. At the finer frequencies, FlowLOB reproduces both the concentration of returns near zero and the small but important spread tail. The bottom row shows that the generated books also recover the main cross-level volume-correlation structure near the touch, as well as the weaker dependence pattern of the less liquid held-out symbol.

% describe Table 1, 0.1s and 1s
We then quantify marginal realism in Table~\ref{tab:realism}, which compares generated and real pooled marginals of the price and volume features. We report Wasserstein-1 ($W_1$), Kolmogorov--Smirnov (KS), Kullback--Leibler (KL), and Jensen--Shannon (JS) distances, with lower values indicating closer agreement. 
% why difflob is absent
We do not include DiffLOB as a separate realism baseline because its setting is not directly matched to our multi-symbol task. Moreover, Sec.~\ref{sec:exp-nfe} has established the efficiency advantage of flow matching by a controlled diffusion--flow comparison.
% Quantification of table 1
At $0.1$s and $1$s, FlowLOB gives the closest marginal match across nearly all comparisons, achieving the best distance in 30 of the 32 price/volume metric cells. Relative to the strongest non-FlowLOB baseline, FlowLOB makes a median $4.5\times$ smaller distance. 
On 700.HK, FlowLOB is best in all cells at $0.1$s and in six of eight cells at $1$s, where ZI slightly improves the price $W_1$ and price KS distances. On the held-out 9999.HK symbol, FlowLOB is best in every $0.1$s and $1$s cell. These gains indicates that the selected low-NFE sampler preserves high-frequency LOB structure. Its strong performance at $0.1$s and $1$s supports the tick-relative representation introduced in Sec.~\ref{sec:data-tensors}: by removing much of the symbol-specific price and volume scale, the same generator can transfer to an unseen instrument without retraining.

% Limitation on 10s frequency
% this can be mitigated by larger universe and longer period
The main limitation appears at the coarsest frequency. At $10$s, FlowLOB no longer dominates the marginal metrics: ZI gives the best price distances on both symbols, and LOB-S5 is competitive or best on several volume distances. The diagnostics in Fig.~\ref{fig:realism-diagnostics} show the same boundary: the generated $10$s return distributions are heavier-tailed than the real distributions, and the volume-correlation plots show stronger cross-level dependence than observed in the data. We interpret this as a limitation of the current model at coarse sampling frequencies, where the same calendar span provides fewer training windows and where returns occupy a wider multi-tick support. The realism advantage of FlowLOB is therefore clearest at the high-frequency settings where fast sampling is most valuable. This limitation may be mitigated by training on a larger symbol universe and a longer historical period.

\subsection{Counterfactual Generation}
\label{sec:exp-counterfactual}

The realism tests above ask whether generated books match market data under realized conditions. We now ask whether the conditioning variables can be used as controls. We use ``counterfactual'' in the simulator sense: the observed past window is held fixed, while one requested scenario variable is replaced by a high- or low-regime value. Controllability is then evaluated on the generated book itself, by recomputing the controlled statistic and checking whether its distribution moves toward the corresponding real tail regime. 
For each scenario variable defined in Sec.~\ref{sec:data-tensors}---trend, volatility, liquidity, and imbalance---and for each direction, high or low, we generate two sets of samples from the same past windows. Reference samples use the realized scenario values. Counterfactual samples replace one scenario variable with values drawn from the corresponding high or low $5\%$ tail of its training-set distribution, while all other conditioning inputs are unchanged. The target regime is the matching high or low $5\%$ tail of the same statistic computed on real test windows.
Let $W_1^{\mathrm{cf}}$ denote the Wasserstein-1 distance between the recomputed statistic of the counterfactual samples and the real test-set tail data, and let $W_1^{\mathrm{ref}}$ denote the corresponding distance for the reference samples. We say the control is valid for a cell when $W_1^{\mathrm{cf}} < W_1^{\mathrm{ref}}$. This criterion is deliberately distributional: it does not reward a model for merely receiving a condition, but only when the generated books themselves exhibit the requested regime more closely than the reference continuation.

% Should explain why there are equal values (especially in high frequency)
Table~\ref{tab:control} shows that conditioning moves generated books toward the requested regime in 42 of 48 cells. 
The most reliable controls are the volume-related variables, liquidity and imbalance, which are valid for all symbols, frequencies, and directions. These variables are computed directly from the generated depth profile at each future step, so their signal is closely aligned with the output tensor. 
For 700.HK, the less reliable cases are concentrated in trend and volatility. Trend shows near-ties at the finest frequency, where the empirical high- and low-trend tail regimes are weakly separated, and its low-regime control fails at $10$s for both symbols.
Volatility-low also fails at $10$s. 
% Why we have counterfactual failure
Thus the main failures occur either when the target statistic is weakly expressed over the generated horizon or at the coarsest frequency, consistent with the realism boundary observed in Sec.~\ref{sec:exp-realism}.
The held-out 9999.HK results are broadly consistent with 700.HK, showing that controllability is not only a symbol-specific effect but also transfers to an unseen symbol.

\begin{table}[t]
\caption{Counterfactual validity: $W_1$ distance from the recomputed
statistic of counterfactual (Cf.) and reference (Ref.) samples to the real $5\%$ tail regime. Bold marks the smaller distance; an asterisk indicates a full-precision win that appears tied after rounding.}
\label{tab:control}
\centering
% \footnotesize
\setlength{\tabcolsep}{3.4pt}
\scalebox{0.87}{
\begin{tabular}{lll rr rr rr}
\toprule
& & & \multicolumn{2}{c}{0.1s} & \multicolumn{2}{c}{1s}
    & \multicolumn{2}{c}{10s} \\
\cmidrule(lr){4-5}\cmidrule(lr){6-7}\cmidrule(lr){8-9}
Symbol & Channel & Sel. & Cf. & Ref. & Cf. & Ref. & Cf. & Ref. \\
\midrule
\multirow{8}{*}{700.HK}
& \multirow{2}{*}{trend} & high & 0.549 & $^*$\textbf{0.549} & \textbf{1.40} & 1.41 & \textbf{5.49} & 6.65 \\
& & low & 0.555 & $^*$\textbf{0.555} & $^*$\textbf{1.37} & 1.37 & 11.5 & \textbf{6.82} \\
& \multirow{2}{*}{volatility} & high & \textbf{0.192} & 0.212 & \textbf{0.151} & 0.323 & \textbf{0.408} & 0.472 \\
& & low & \textbf{0.006} & 0.008 & \textbf{0.042} & 0.067 & 0.874 & \textbf{0.866} \\
& \multirow{2}{*}{liquidity} & high & \textbf{0.205} & 0.463 & \textbf{0.217} & 0.505 & \textbf{0.619} & 0.807 \\
& & low & \textbf{0.404} & 0.580 & \textbf{0.255} & 0.539 & \textbf{0.251} & 0.288 \\
& \multirow{2}{*}{imbalance} & high & \textbf{0.517} & 0.551 & \textbf{0.228} & 0.562 & \textbf{0.340} & 0.450 \\
& & low & \textbf{0.327} & 0.338 & \textbf{0.092} & 0.326 & \textbf{0.405} & 0.440 \\
\midrule
\multirow{8}{*}{9999.HK}
& \multirow{2}{*}{trend} & high & $^*$\textbf{0.536} & 0.536 & $^*$\textbf{1.61} & 1.61 & \textbf{6.34} & 6.66 \\
& & low & $^*$\textbf{0.534} & 0.534 & \textbf{1.65} & 1.66 & 9.41 & \textbf{6.14} \\
& \multirow{2}{*}{volatility} & high & \textbf{0.186} & 0.196 & \textbf{0.163} & 0.366 & \textbf{0.632} & 0.658 \\
& & low & \textbf{0.006} & 0.008 & \textbf{0.043} & 0.071 & 0.827 & \textbf{0.817} \\
& \multirow{2}{*}{liquidity} & high & \textbf{0.187} & 0.278 & \textbf{0.047} & 0.323 & \textbf{0.552} & 0.585 \\
& & low & \textbf{0.520} & 0.592 & \textbf{0.135} & 0.547 & \textbf{0.302} & 0.305 \\
& \multirow{2}{*}{imbalance} & high & \textbf{0.273} & 0.283 & \textbf{0.081} & 0.303 & \textbf{0.247} & 0.279 \\
& & low & \textbf{0.306} & 0.318 & \textbf{0.059} & 0.298 & \textbf{0.288} & 0.325 \\
\bottomrule
\end{tabular}}
\end{table}

\subsection{Ablation Study}
\label{sec:exp-ablation}

\begin{figure*}[t]
\centering
\includegraphics[width=0.9\textwidth]{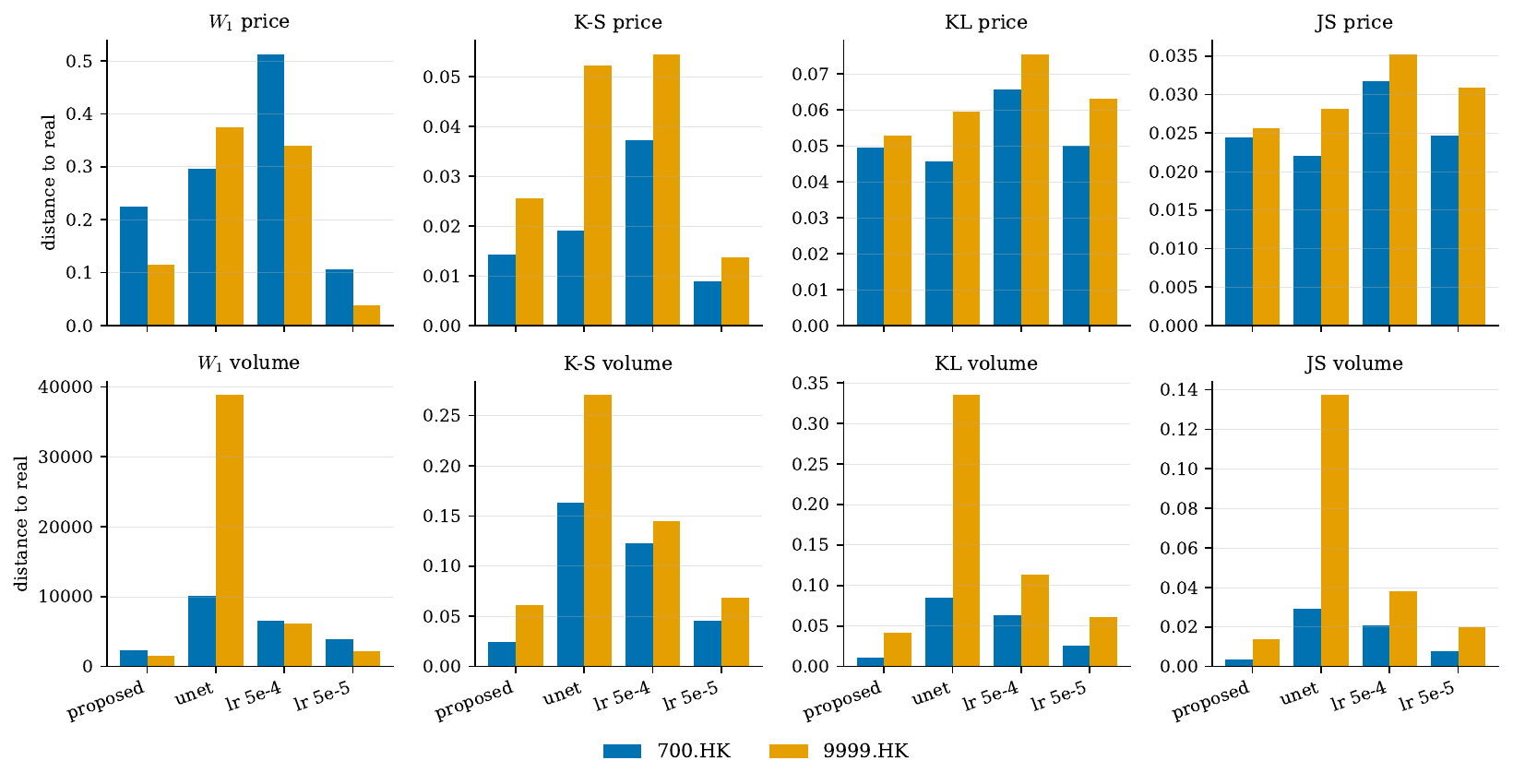}
\caption{Ablation results at $1$s. We compare the proposed transformer
configuration against a parameter-matched UNet and two learning-rate variants. Bars show distances between
generated and real pooled price and volume marginals for the in-distribution
symbol 700.HK and the held-out symbol 9999.HK; lower is better.}
\label{fig:ablations}
\end{figure*}

We finally ablate the two design choices likely to affect the preceding claims: the transformer backbone and the learning rate. The ablation is not intended to exhaust the architecture space; rather, it tests whether the proposed configuration is robust under changes that directly affect realism and zero-shot transfer. For the backbone ablation, we replace the transformer with a UNet of comparable parameter count while keeping the data representation and sampler fixed. For the learning-rate ablation, we compare the default rate $1e-4$ with one higher and one lower rate. All ablations are run at the $1$s frequency, sampled with Euler at $N=10$, and evaluated with the same marginal distances as in Sec.~\ref{sec:exp-realism}.

Figure~\ref{fig:ablations} shows that the transformer backbone is important, especially out of distribution. A parameter-matched UNet trails the proposed model on most price and volume metrics, and the degradation is largest on the held-out 9999.HK symbol. This suggests that attending over the level--time grid helps the model use the shared tick-relative representation for cross-symbol generalization, rather than merely fitting the in-distribution symbol.

The learning-rate ablation shows a smaller but useful trade-off. A higher learning rate degrades both price and volume realism, while a lower learning rate improves some price marginal distances but weakens volume fidelity. We therefore keep the default learning rate as a balanced operating point rather than selecting a configuration tuned for a single metric. Across the ablations, the qualitative ranking is similar on 700.HK and 9999.HK, supporting the conclusion that the realism and controllability results above are not artifacts of a lucky training setting.

\section{Conclusion}
\label{sec:conclusion}

We introduced FlowLOB, a conditional flow-matching model for generating limit order book trajectories. The model combines a tick-relative LOB representation, scenario conditioning, and a level--time transformer backbone to train a single generator across multiple HKEX symbols. Under a matched comparison with diffusion, flow matching reaches its best sample quality with only 10 Euler steps, giving a practical low-NFE operating point for generating many market trajectories across symbols and regimes. At this efficient operating point, FlowLOB matches high-frequency price and volume distributions more closely than the evaluated agent-based and learned baselines on most metrics, transfers to a held-out symbol without retraining, and supports measurable control over liquidity and imbalance regimes.

The results should be read as a first step rather than a final account of market simulation. The current model is strongest at the finer sampling frequencies, while its advantage weakens at coarser resolution; trend control is also less reliable over short windows. These limitations are useful because they identify where larger data, broader symbol coverage, longer contexts, and different conditioning designs should matter most. More broadly, this paper provides the experimental basis for studying scaling laws in financial market simulation. FlowLOB gives a controlled setting in which data scale, model size, sampling cost, symbol diversity, and controllability can be varied systematically. The next step is to move from showing that flow-based LOB generation is efficient, realistic, and steerable in this setting to quantifying how these properties improve as market simulators are scaled with bigger models and more budgets.

\bibliographystyle{ACM-Reference-Format}
\bibliography{references}

\end{document}